\documentclass[german,english]{bvm2013}
\RequirePackage{wrapfig}

\title{Co-Registration of Intra-Operative Photographs and Pre-Operative MR Images}
\titlerunning{Registration of Intra-Operative Photographs and Pre-Operative MR Images}

\author{Benjamin Berkels$^1$, Ivan Cabrilo$^2$, Sven Haller$^2$,\\Martin Rumpf$^1$, Carlo Schaller$^2$}

\authorrunning{Berkels et al.}

\institute{%
$^1$Institut f\"ur Numerische Simulation, Rheinische Friedrich-Wilhelms-Universit\"at Bonn\\
$^2$H\^{o}pitaux Universitaires de Gen\`{e}ve}

\email{benjamin.berkels@ins.uni-bonn.de}

\begin{document}

\selectlanguage{english}

\maketitle

\begin{abstract}
Brain shift, i.\thinspace e. the change in configuration of the brain after opening the dura mater, is a key problem in neuronavigation. We present an approach to co-register intra-operative microscope images with pre-operative MRI data to adapt and optimize intra-operative neuronavigation.
The tools are a robust classification of sulci on MRI extracted cortical surfaces,
guided user marking of most prominent sulci on a microscope image, and the actual variational registration method with a fidelity energy for 3D deformations of the cortical surface combined with a higher order, linear elastica type prior energy.
Furthermore, the actual registration is validated on an artificial testbed and on real data of a neuro clinical patient.
\end{abstract}

\section{Introduction}
The development of medical imaging in the last decades quickly triggered intense interest from the medical world to translate this progress on the imaging side to clinical diagnostics and treatment planning.
In that respect image registration and in particular recently also the fusion of 2D and 3D image data is a fundamental task in image--guided medical intervention.
In \cite{2003-01} 2D photographs of human faces are registered with a triangulated facial surface extracted from MRI data using rigid deformations.
A registration method for sparse but highly accurate 3-D line measurements with a surface extracted from volumetric planning data based on the consistent registration idea and higher order regularization is introduced in \cite{2003-02}.

The matching of photographic images with pre-operative MRI data is a particular challenge in
cranial neuronavigation.
The photograph--MRI registration problem in the context of intracranial electroencephalography
has been investigated via a control point matching approach in \cite{2003-03}.
Recently, normalized mutual information
has been applied for the rigid transformation co-registration of brain photographs and MRI extracted cortical surfaces \cite{2003-04}.
A major limitation of note, however, is that due to the brain shift the surgeon's view of the operating site is not in a rigid transformation correspondence to pre-operative images.
Indeed, standard intracranial neuronavigation devices do not correct for this movement of brain \cite{2003-05}.
The main contributions of this paper are a novel classification method for crease pattern
such as sulci on implicit (cortical) surfaces and the actual 2D-3D registration method, where a non-rigid 3D
deformation of the cortical surface is identified based on user marked sulci on photographs and the camera
parameters.

\section{Materials and Methods}
\begin{wrapfigure}[8]{r}{0.6\linewidth}
\centering
\begin{tabular}{cc}
\\[-7ex]
\includegraphics[height=0.31\linewidth]{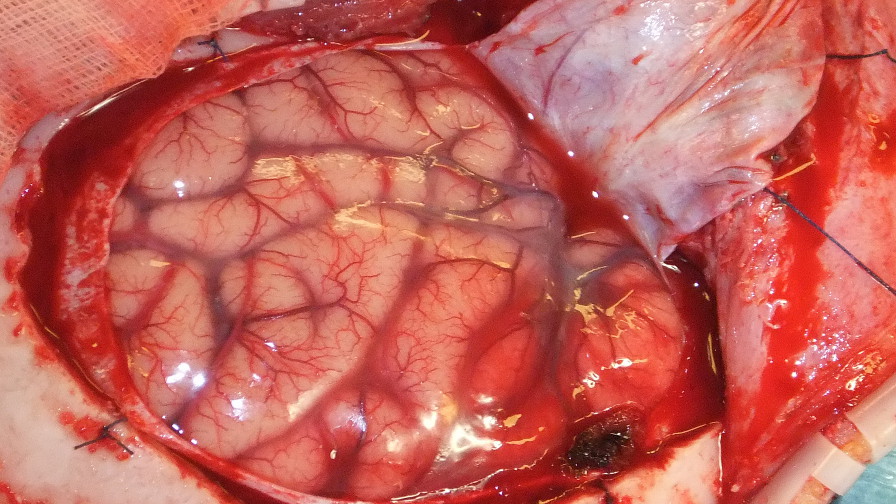}&\includegraphics[height=0.31\linewidth]{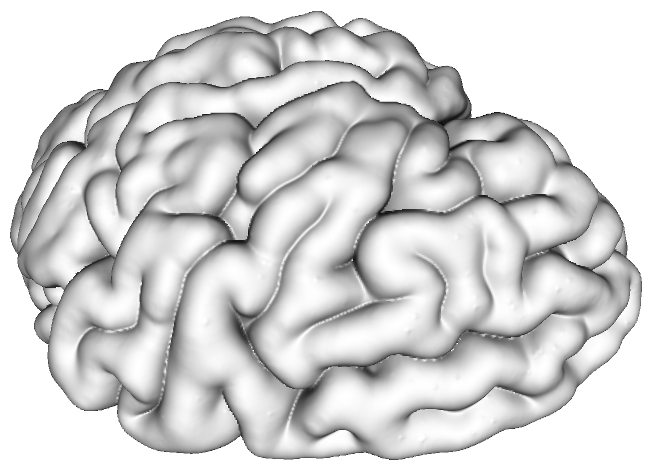}\\[-2ex]
\end{tabular}
\caption{Input photograph, MRI graph surface}
\label{fig:inputData}
\end{wrapfigure}

The aim of this paper is to register a photograph of the exposed human cortex with the cortex geometry extracted from an MRI data set, see Fig.~\ref{fig:inputData}, using the sulci as fiducials. The main ingredients of the proposed approach are a sulci classification on the cortex geometry (Section~\ref{sec:GeomClassification}) and on the photograph (Section~\ref{sec:PhotoClassification}),
as well as a model that uses the two classifications to register
photograph and cortical geometry (Section~\ref{sec:RegisModel}).
The 2D digital photographs were taken intra-operatively after supratentorial craniotomy and durotomy, and before corticotomy, using a digital camera with 10 mega pixel resolution positioned 20 cm above the craniotomy.
The MR imaging was performed on a 3T MRI scanner. A T1-weighted MP2RAGE sequence ($1\times1\times1\;m^3$
, $256\times256\times176$ matrix)
was segmented into gray and white matter using BrainVoyager QX \cite{2003-06} and then converted into a signed distance function of the cortical surface using a fast marching method.

\subsection{Sulci Classification on MRI Data}
\label{sec:GeomClassification}
In this section, we describe how to classify creases on the contour surface of a 3D
object $\mathcal{B}\subset\Omega$ represented via its signed
distance function $d:\Omega\rightarrow\mathbb{R}$ on a computational domain $\Omega$.
In the application the object is a brain volume
\begin{wrapfigure}[8]{r}{0.7\linewidth}
\centering
\begin{tabular}{cccc}
\\[-7ex]
\includegraphics[width=0.24\linewidth]{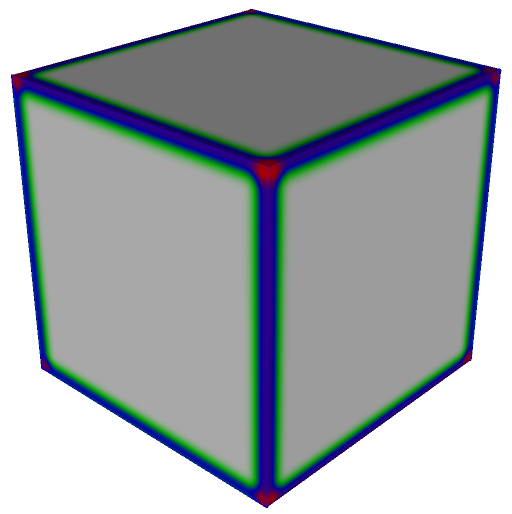}&
\includegraphics[width=0.24\linewidth]{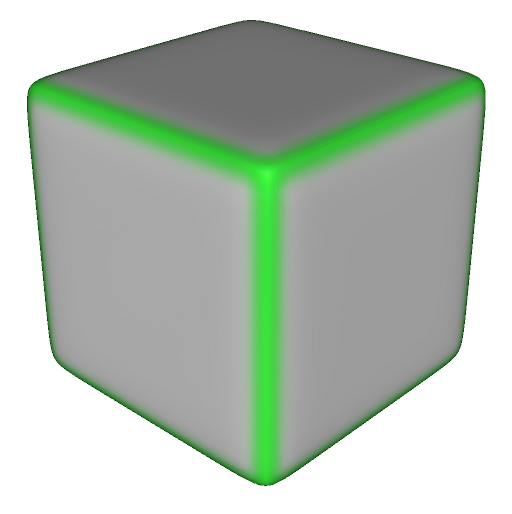}&
\includegraphics[width=0.24\linewidth]{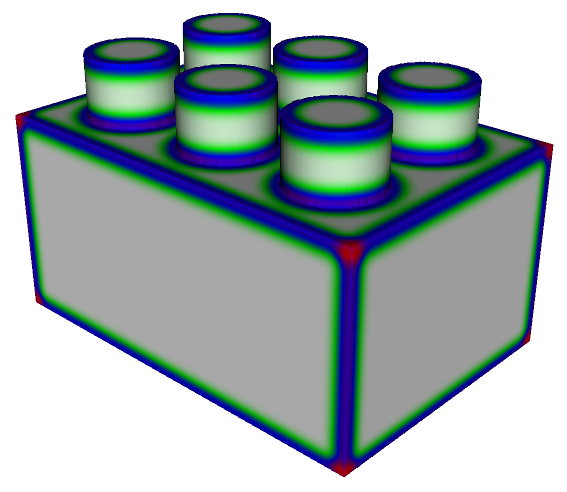}&
\includegraphics[width=0.24\linewidth]{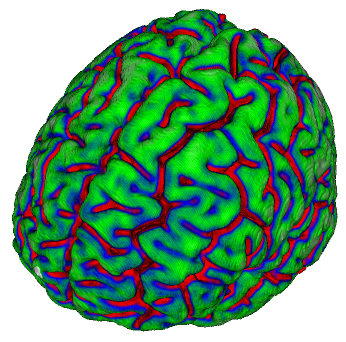}\\[-2ex]
\end{tabular}
\caption{Moment-based classification}
\label{fig:MomentClassification}
\end{wrapfigure}
and the creases are the sulci on the cortical surface.
We aim for a moment based analysis of the (cortical) surface $\mathcal{C}:=\partial \mathcal{B}$ and
define the zero moment shift of the implicit surface $\mathcal{C}$ as follows
$M_{\epsilon}^0[\mathcal{B}](x)=\frac{1}{|B_\epsilon(x)|}\int_{B_\epsilon(x)}d(y)(y-x)\mathrm{d}y$ which returns larger values
in flat regions of $\mathcal{C}$ than in edge regions and even smaller near corners (cf. \cite{2003-07}
for a related moment based classification on explicit surfaces).
We define the scalar classification $\mathbf{C}(x)=g_{\beta}\left( { \lVert M_{\epsilon}^0[\mathcal{B}](x)} \rVert/{ \epsilon^2 } \right)$,
where $g_{\beta}(t)=\frac{1}{1 + \beta t^2}$.
Fig.~\ref{fig:MomentClassification} illustrates the behavior of the classifier $\mathbf{C}$ on three simple shapes and a cortical surface extracted from an MRI using a white-green-blue-red color coding. We observe a robust distinction for a single set of parameters ($\beta=20$ and $\epsilon=8h$ or $4h$, where $h$ denotes the grid width).
\subsection{Generation of Annotated Cortex Photographs}
\label{sec:PhotoClassification}
\begin{wrapfigure}[8]{r}{0.45\linewidth}
\centering
\begin{tabular}{cc}
\\[-8ex]
\includegraphics[width=0.46\linewidth]{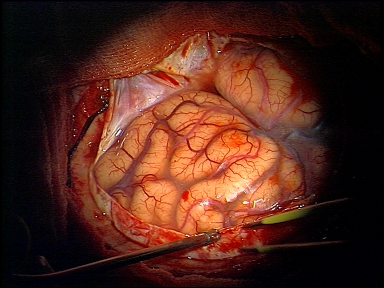}&
\includegraphics[width=0.46\linewidth]{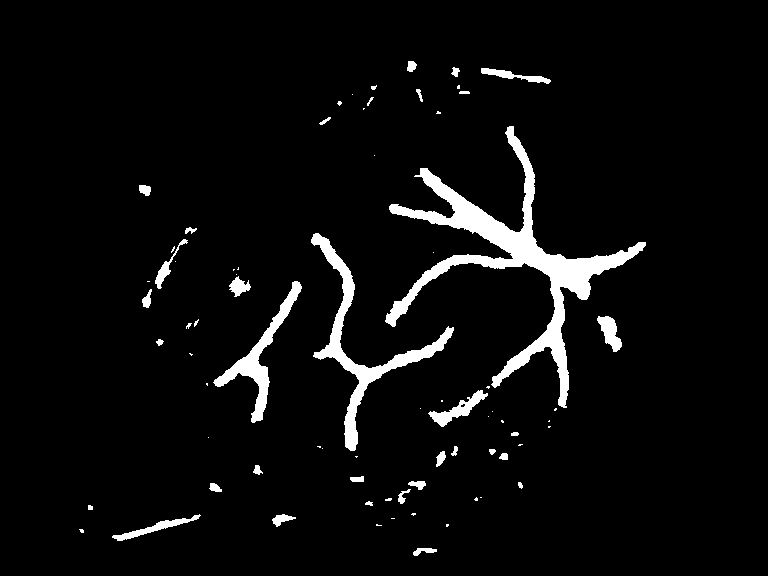}\\[-2ex]
\end{tabular}
\caption{Cortex photograph and dictionary based pre-classification}
\label{fig:DictSegmentation}
\end{wrapfigure}
Essential problems for the classification of sulci on photographs are additional structures and their misinterpretation. Most prominent are cortical veins,
 which in addition partially occlude sulci
(these veins are almost invisible in MRI in the used MP2RAGE sequence).
On this background we here confine to a still manual marking of
sulci by an expert who is supported by
the results of a prior automatic pre-classification of sulci based on learned discriminative dictionaries,  cf. Fig.~\ref{fig:DictSegmentation}, where the method from \cite{2003-08} is used.

\subsection{Registration of Photograph and Cortex Geometry}
\label{sec:RegisModel}
The co-registration of an MRI extracted cortical surface $\mathcal{C}\subset\mathbb{R}^3$ and a
photograph to compensate for effects such as the brain shift is based on a co-registration of the sulci classifiers on $\mathcal{C}$  and the photograph. To this end, we
\begin{wrapfigure}[12]{r}{0.45\linewidth}
\centering
\begin{tabular}{c}
\\[-7.5ex]
\includegraphics[width=\linewidth]{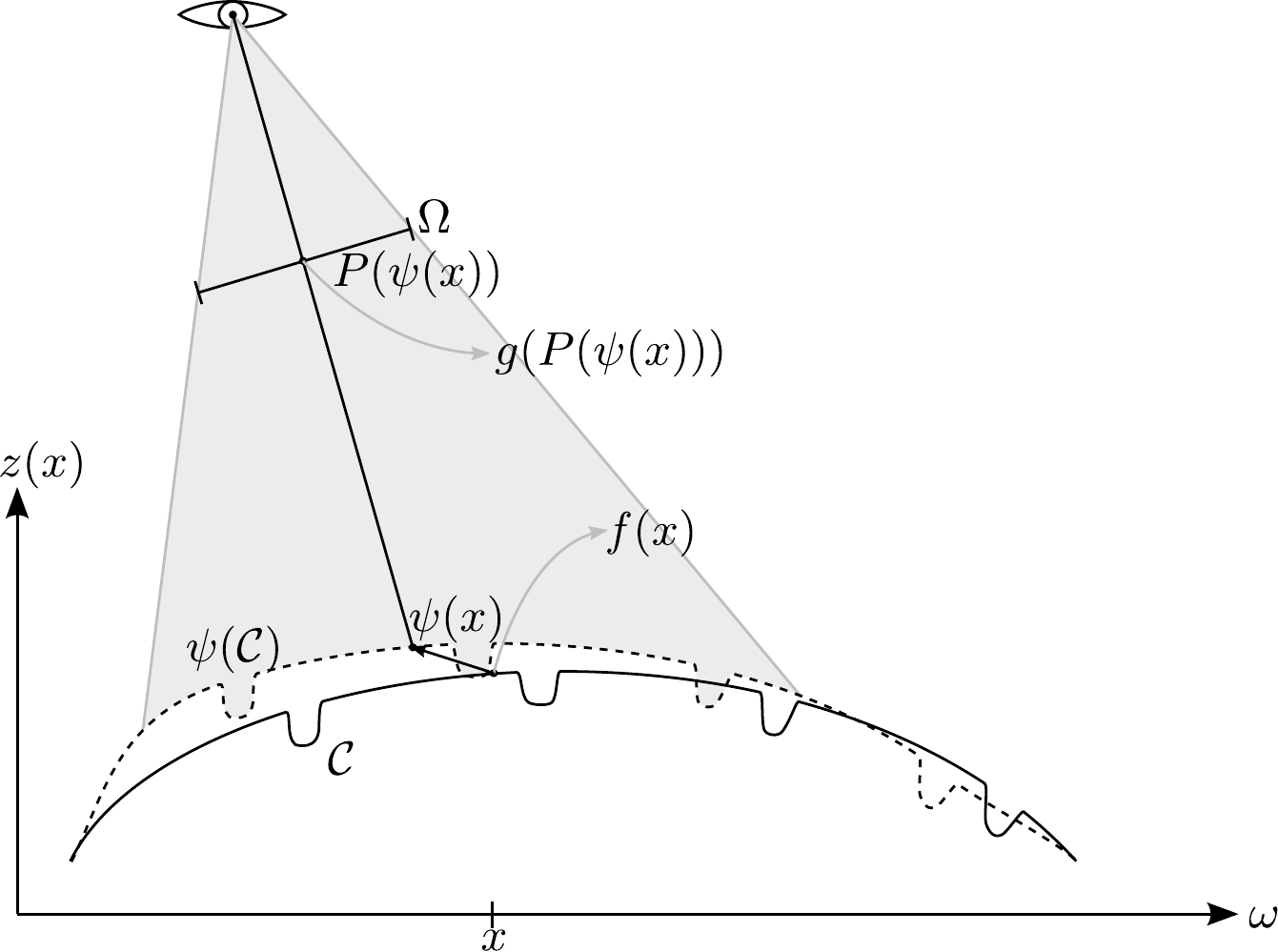}\\[-3ex]
\end{tabular}
\caption{Geometric configuration}
\label{fig:ConfigSketch}
\end{wrapfigure}
 suppose $\mathcal{C}$ to be described as a graph
\[\mathcal{C}=\{(x,z(x))\in\mathbb{R}^3\vert x\in\omega \}\] with parameter domain
$\omega\subset\mathbb{R}^2$ and graph function $z:\omega \to \mathbb{R}$.
We are interested in  a local registration described by the craniotomy,
where such a graph representation can be easily
derived from the signed distance function used in Section~\ref{sec:GeomClassification}.
Furthermore, let $g \in L^2(\Omega)$ denote the sulci classifier on the photograph domain $\Omega$ (cf. Section \ref{sec:PhotoClassification}),
and $f\in L^2(\omega)$ the corresponding classifier on the cortical surface given as a function on the parameter domain $\omega$, obtained by a suitable clamping of $\mathbf{C}$ and rescaling of the values to the unit interval $[0,1]$ (cf. Section~\ref{sec:GeomClassification}).
Both classifiers are supposed to be close to $1$ in the central region of the sulci and small outside.
Finally, we denote by $P:\mathbb{R}^3\rightarrow\Omega$ the  projection of points in $\mathbb{R}^3$ onto the image plane $\Omega$ derived from known camera parameters.
Let us remark that we thereby implicitly rule out self occlusions of the graph surface $\mathcal{C}$ under the image plane projections.
Now, we ask for a deformation
$\Psi:\omega \rightarrow\mathbb{R}^3$ defined on the parameter domain $\omega$ of the graph function $z$ that matches $\mathcal{C}$ to its deformed configuration represented under the projection $P$  in the photograph. Thereby, matching is encoded via the coincidence of the surface classifier $f(x)$ on the MRI described cortical surface and the image classifier $g(P(\psi(x)))$ evaluated at the projected deformed position $P(\psi(x))$ for all $x$ on the parameter domain $\omega$.
Thus, proper matching can be encoded via the minimization of the matching energy
\begin{equation*}
E_{\mbox{\tiny match}}[\psi]=\frac{1}{2}\int_\omega[g(P(\psi(x)))-f(x)]^2
A(x) \mathrm{d}x
\end{equation*}
based on a surface integral over $\mathcal{C}$ with the area element
$A(x) = (1+|\nabla z(x)|^2)^{\frac12}$, to consistently reflect the cortex geometry.
In the overall variational approach the matching energy is complemented by a suitable elastic regularization energy, which acts as a prior on admissible deformations $\psi$.
Here, we consider the second order, elastic energy
\begin{equation*}
E_{\mbox{\tiny reg}}[\psi]
=\frac{1}{2}\int_\omega |\Delta \psi_1(x)|^2+|\Delta \psi_2(x)|^2+|\Delta \psi_3(x)-\Delta z(x))|^2 \mathrm{d}x.
\end{equation*}
Note that a simple first order regularization like the Dirichlet energy of the displacement \mbox{$\psi - (\cdot, z(\cdot))$}
is not sufficient since matching information is mostly given on a low dimensional subset where proper nonlinear extrapolation is required and bending modes play an important role. Obviously, $E_{\mbox{\tiny reg}}$ is rigid body motion invariant (cf. \cite{2003-09}).
Finally, we combine the matching energy $E_{\mbox{\tiny match}}$ and the regularization energy $E_{\mbox{\tiny reg}}$ to the total energy functional
$E[\psi] = E_{\mbox{\tiny match}}[\psi]+\lambda E_{\mbox{\tiny reg}}[\psi]$
on deformations $\psi$ encoding the deformation of the
cortical surface $\mathcal{C}$, where $\lambda$ is a positive constant controlling the strength of the regularization.

To minimize the objective functional we use a time discrete regularized gradient descent taking into account a suitable step size control combined with a cascadic descent approach to handle the registration in a coarse to fine manner. The first variation necessary for the descent algorithm is
\begin{align*}
\left<E^\prime[\psi],\zeta\right>=&\int_\omega[g(P(\psi(x)))-f(x)]\nabla g(P(\psi(x)))\cdot DP(\psi(x))
 \zeta(x) A(x) \mathrm{d}x\\
& +\lambda \int_\omega (\Delta^2 \psi_1,\Delta^2 \psi_2,\Delta^2 (\psi_3 - z)) \cdot  \zeta \mathrm{d}x,
\end{align*}
where the natural boundary conditions $\partial_\nu \Delta\psi = \Delta\psi =0$ on $\partial \omega$ for the normal $\nu$ on $\partial \omega$ are considered and  $\psi$ is initialized as the identity on $\mathcal{C}$, i.\thinspace e.
\mbox{$\psi(x) = (x,z(x))$}. For the spatial discretization we consider bilinear Finite Elements on a rectangular mesh overlaying $\omega$ and $\Omega$, and approximate the bi-Laplacian $\Delta^2$ by the squared standard discrete Laplacian $\Delta^2_h=M^{-1}L M^{-1}L$. Here, $M$ and $L$ denote the standard (lumped) mass and stiffness matrices, respectively.

\section{Results}
\begin{figure}[t]
\centering
\begin{tabular}{cccc}
\includegraphics[height=0.17\linewidth]{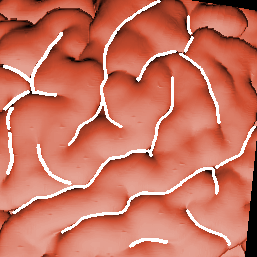} &
\includegraphics[height=0.17\linewidth]{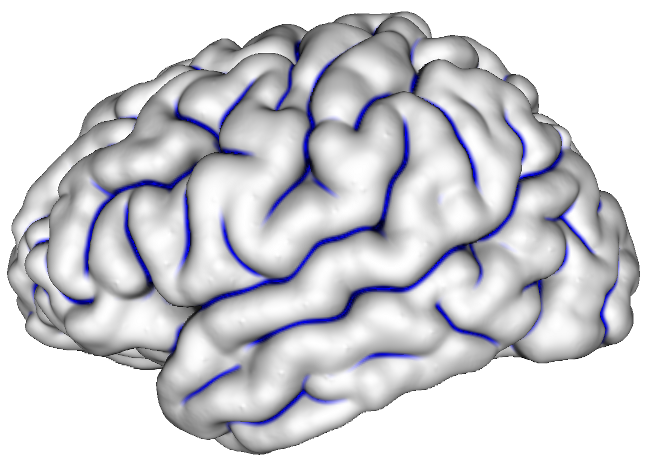} &
\includegraphics[height=0.17\linewidth]{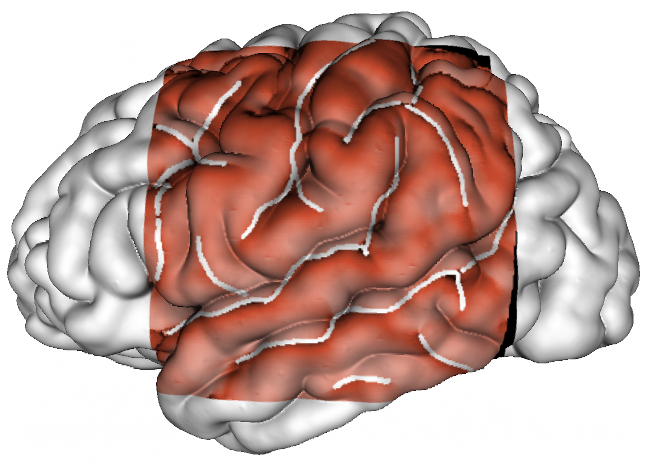} &
\includegraphics[height=0.17\linewidth]{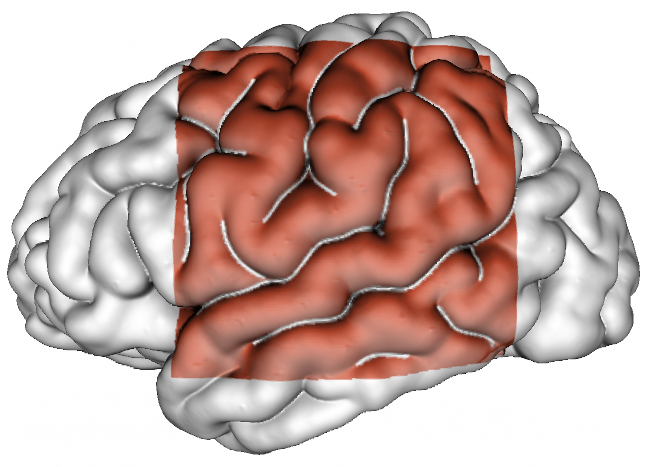}\\[1ex]
\hline
\\
\includegraphics[height=0.17\linewidth]{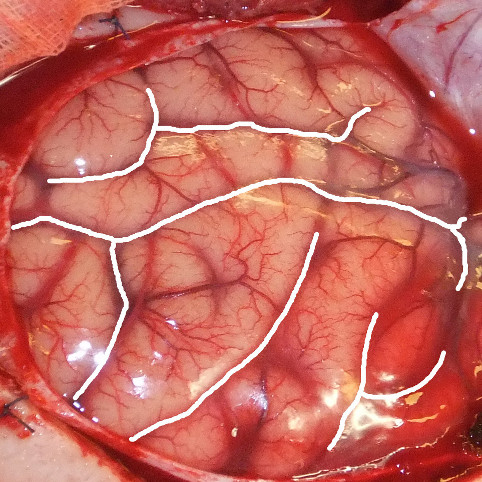} &
\includegraphics[height=0.17\linewidth]{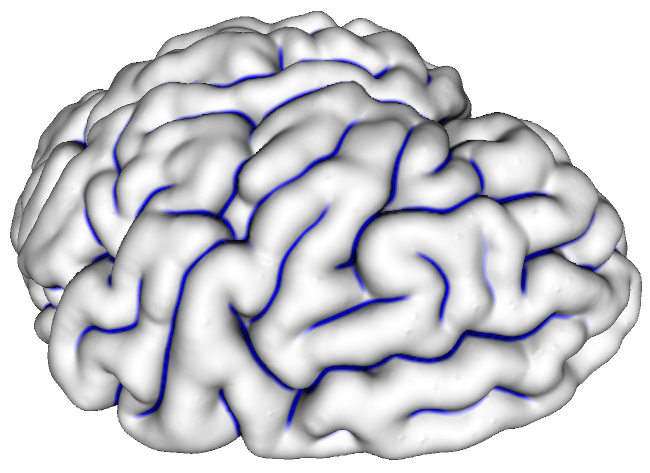} &
\includegraphics[height=0.17\linewidth]{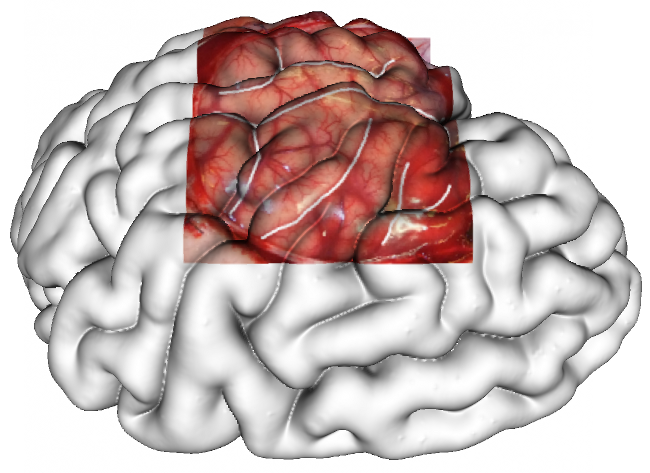} &
\includegraphics[height=0.17\linewidth]{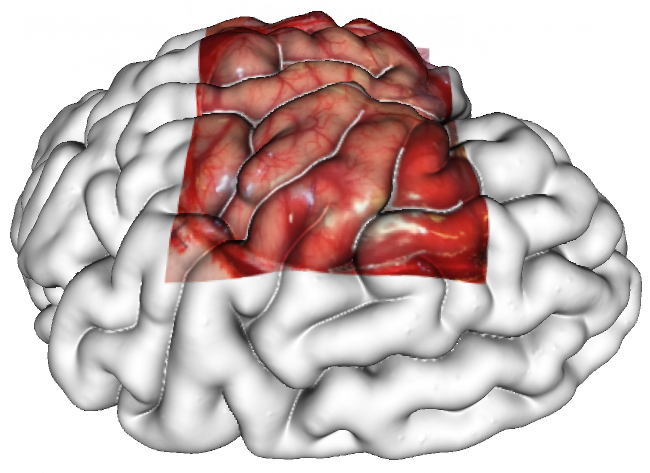}\\
\begin{minipage}{0.22\linewidth}\centering manual segmentation\end{minipage}&
\begin{minipage}{0.22\linewidth}\centering moment-based segmentation\end{minipage}&
\begin{minipage}{0.22\linewidth}\centering before registration\end{minipage}&
\begin{minipage}{0.22\linewidth}\centering after registration\end{minipage} \\
\end{tabular}
\caption{\label{fig:res} Both for the test data (top row) and the pair of  a true photograph and an MRI extracted cortical surface (bottom row) we show (from left to right) the input image with marked sulci (computed image and real photograph respectively), the sulci classification on the 3D cortical surface segmented from MRI data, the initial misfit of the sulci marking on the 2D image projected on the cortical surface overlaying the cortical surface itself, and the final registration result.}
\end{figure}

We have applied our registration approach both to test data and to real data. For the test data a cortical surface segmented on a 3D MRI data set has been taken as input together with an image generated from this surface via a given projection concatenated with
an additional 3D nonrigid deformation.
Then on the projected image selected sulci have been marked by hand.
This pair of input data together with the computed registration result are depicted in Fig.~\ref{fig:res}.

Furthermore, we considered a photograph of the brain surface seen through a left fronto-temporo-parietal craniotomy performed in a patient before placement of a subdural electrode grid for investigation of drug-resistant cryptogenic epilepsy.
We segmented sulci using the procedure in Section \ref{sec:PhotoClassification}.
This is then registered via the proposed approach with the cortical surface extracted from the corresponding pre--interventional 3D MRI data.
Again input data and achieved registration are shown in Fig.~\ref{fig:res}.
\section{Discussion}
\label{sec:discussion}
We have proposed a novel method for the registration of photographs (2D) of the brain with the cortical surface
extracted from 3D MRI data. The method turns out to be effective and robust both on test and on real data. It can be considered as an alternative to intra-operative MRI allowing subsequent co-registration with neuronavigation \cite{2003-10}.  Currently, we aim for a validation study with an increased number of cases considering also data of patients with substantially smaller craniotomies.

Furthermore, there is potential, that the iterative 2D/3D surface registration of digital images together with morphological 3D MRI data sets will enable to build up a ``dictionary'' of brain surface features. Ultimately, the creation of such a dictionary might, to a certain extent, permit ``intelligent'' automatic recognition of brain surface features, where the 2D brain surface, seen through the intra-operative microscope standardly used during intra-cerebral procedures, would directly be co-registered to pre--interventional 3D MRI data.

As already discussed, the sensitivity of MRI and photography is substantially different for different anatomic structures, e.\thinspace g. veins are very prominent on images yet not on the MRI modality used here.
One could incorporate multiple MRI sequences and fuse vein sensitive images with the present images to improve the registration results.
Finally, let us remark that one could also consider stereo photographs of the deformed surface
to improve the methods performance. In that respect our approach can easily be adapted
summing over copies of the matching energy.

\bibliographystyle{bvm2013}

\bibliography{2003}

\end{document}